\documentclass[11pt,a4paper]{article}
\usepackage[breaklinks]{hyperref}
\usepackage[hyphenbreaks]{breakurl}
\usepackage[hyperref]{acl2019}
\usepackage[utf8]{inputenc}
\usepackage{times}
\usepackage{latexsym}
\usepackage{multirow}
\usepackage{array}
\usepackage{graphicx}
\usepackage{arydshln} 
\usepackage[small]{titlesec} 
\usepackage{url}
\usepackage{amsmath}
\usepackage{microtype}
\usepackage[absolute]{textpos}

\newcommand{\squishlist}{
 \begin{list}{$\bullet$}
  { \setlength{\itemsep}{0pt}
     \setlength{\parsep}{2pt}
     \setlength{\topsep}{2pt}
     \setlength{\partopsep}{0pt}
     \setlength{\leftmargin}{1em}
     \setlength{\labelwidth}{1em}
     \setlength{\labelsep}{0.4em} } }

\newcommand{\squishend}{
  \end{list}  }

\aclfinalcopy 
\def\aclpaperid{122} 

\newcommand\Tstrut{\rule{0pt}{2.3ex}}       

\definecolor{darkgreen}{rgb}{0.0, 0.5, 0.0}
 
 \newcommand{\OD}[1]{\textbf{\color{darkgreen} (OD: #1)}}
 
\usepackage[normalem]{ulem}
\def\ODdel#1{\bgroup\markoverwith{\textcolor{darkgreen}{\rule[0.5ex]{2pt}{1pt}}}\ULon{#1}}
\def\davedel#1{\bgroup\markoverwith{\textcolor{blue}{\rule[0.5ex]{2pt}{1pt}}}\ULon{#1}}

\newcommand{\rnnlg}{\textsc{RNNLG}}

\title{Improved Semantics for the End-to-End Generation Challenge Corpus}

\title{Noise Matters: Cleaning the E2E challenge dataset}

\title{How \OD{Training Data} Noise influences Neural Generation}

\title{How training data with noisy semantics influences neural NLG}

\title{Semantic Noise Matters for Neural NLG}
\title{Semantic Noise Matters for Neural Natural Language Generation}

\author{Ondřej Dušek$^\ast$\\ 
Charles University\\ 
Faculty of Mathematics and Physics \\
Prague, Czech Republic\\
\hspace{-2em}\texttt{odusek@ufal.mff.cuni.cz}
\And 
David M. Howcroft$^\ast$ {\rm \&} Verena Rieser\\
The Interaction Lab, MACS\\
Heriot-Watt University \\
Edinburgh, Scotland, UK\\
\texttt{\{d.howcroft,v.t.rieser\}@hw.ac.uk}}

\date{}

\begin{document}
\maketitle

\begin{textblock*}{\textwidth}(2.5cm,1cm)
In \emph{Proceedings of INLG}, Tokyo, Japan, October 2019.
\end{textblock*}

\begin{abstract}

\setcounter{footnote}{1}
\renewcommand{\thefootnote}{\fnsymbol{footnote}}
\footnotetext{Denotes equal contribution.}
\setcounter{footnote}{0}

Neural natural language generation (NNLG) systems are known for their pathological outputs, i.e.\ generating text which is unrelated to the input specification. 
In this paper, we show the impact of semantic noise on state-of-the-art NNLG models which implement different semantic control mechanisms.
We find that cleaned data can improve semantic correctness by up to 97\%, while maintaining fluency. 
We also find that the most common error is omitting information, rather than hallucination.
 
\end{abstract}

\section{Introduction}
\label{sec:introduction}

Neural Natural Language Generation (NNLG) is promising for generating text from Meaning Representations (MRs) in an  `end-to-end' fashion, i.e.\ without needing alignments \cite{wen:emnlp2015, Wen:NAACL16, Dusek:ACL16, Mei:NAACL2016}. 
However, NNLG requires large volumes of in-domain data, which is typically crowdsourced \citep[e.g.][]{mairesse:acl2010,novikova:INLG2016,wen:emnlp2015,Wen:NAACL16,howcroft.etal2017:interspeech}, introducing noise. 
For example, up to 40\% of the E2E Generation Challenge\footnote{\url{http://www.macs.hw.ac.uk/InteractionLab/E2E/}} data contains omitted or additional information \cite{dusek_evaluating_2019}. 

In this paper, we examine the impact of this type of semantic noise on two state-of-the-art NNLG models with different semantic control mechanisms: TGen \cite{Dusek:ACL16} and SC-LSTM \cite{wen:emnlp2015}.
In particular, we investigate the systems' ability to produce fact-accurate text, i.e.\ without omitting or hallucinating information, in the presence of semantic noise.\footnote{Also see \url{https://ehudreiter.com/2018/11/12/hallucination-in-neural-nlg/}}
We find that:
\squishlist
    \item training on cleaned data reduces slot-error rate up to 97\% on the original evaluation data;
    \item testing on cleaned data is challenging, even for models trained on cleaned data, likely due to increased MR diversity in the cleaned dataset; and
    \item TGen performs better than SC-LSTM, even when cleaner training data is available. We hypothesise that this is due to differences in how the two systems handle semantic input and the degree of delexicalization that they expect.
\squishend

In addition, 
we release our code and a cleaned version of the E2E data with this paper.\footnote{Data cleaning scripts, the resulting cleaned data and links to code are available at \url{https://github.com/tuetschek/e2e-cleaning}.}

\section{Mismatched Semantics in E2E Data}
\label{sec:mismatched-semantics}

The E2E dataset contains input MRs and corresponding target human-authored textual references in the restaurant domain. 
MRs here are sets of attribute-value pairs (see Figure~\ref{fig:e2e-data}). 
Most MRs in the dataset have multiple references (8.1 on average).
These were collected using crowdsourcing, leading to noise when crowd workers did not verbalise all attributes or added information not present in the MR. 
According to \citet{dusek_evaluating_2019}, the multiple references should help NLG systems abstract from the noise. 
However, most NLG systems in the E2E challenge in fact produced noisy outputs, suggesting 
 that they were unable to learn to ignore noise in the training input.

\begin{figure}[tb]
\scriptsize\raggedright
\textbf{Original MR}: name[Cotto], eatType[coffee shop], food[English], priceRange[less than £20], customer\_rating[low], area[riverside], near[The Portland Arms]

\medskip
\textbf{Human reference 1 (accurate):} At the riverside near The Portland Arms, Cotto is a coffee shop that serves English food at less than £20 and has low customer rating.

\medskip
\textbf{HR 2:} Located near The Portland Arms in riverside, the Cotto coffee shop serves English food with a price range of £20 and a low customer rating.\\
\textbf{Corrected MR:} name[Cotto], eatType[coffee shop], food[English], customer\_rating[low], area[riverside], near[The Portland Arms] \\\emph{(removed price range)}

\medskip
\textbf{HR 3:} Cotto is a coffee shop that serves English food in the city centre.  They are located near the Portland Arms and are low rated.\\
\textbf{Corrected MR:} name[Cotto], eatType[coffee shop], food[English], customer\_rating[low], area[city centre], near[The Portland Arms] \\\emph{(removed price range, changed area)}

\medskip
\textbf{HR 4:} Cotto is a cheap coffee shop with one-star located near The Portland Arms.\\
\textbf{Corrected MR:} name[Cotto], eatType[coffee shop], priceRange[less than £20], customer rating[low], near[The Portland Arms] \\\emph{(removed area)}
\caption{MR and references from the E2E corpus. The first reference is accurate and verbalises all attributes, but the remaining ones contain inaccuracies. Corrected MRs were automatically produced by our slot matching script (see Section~\ref{sec:cleaning-mrs}). Note that HR 2 is not fixed properly since the script's patterns are not perfect.}
\label{fig:e2e-data}
\end{figure}

Problems with the semantic accuracy in training data is not unique to the E2E dataset.
\citet{howcroft.etal2017:interspeech} collected a corpus of paraphrases differing with respect to information density for use in training NLG systems and found that subjects' paraphrases dropped about 5\% of the slot-value pairs from the original texts and changed the values for approximately 10\% of the slot-value pairs.
As a result of these changes and the insertion of new facts, only 61\% of the corpus contained all and only the intended propositions.
This is similar to what \citet{eric.etal2019} found in their work on the MultiWOZ 2.0 dataset: correcting the dialogue state annotations resulted in changes to about 40\% of the dialogue turns in their dataset.
These findings suggest that efforts to create more accurate training data---whether through stricter crowdsourcing protocols, conducting follow-up annotations (cf.\ \citealp{eric.etal2019}), or automated cleanup heuristics like we report here---are likely necessary in the NLG and dialogue systems communities.

\section{Cleaning the Meaning Representations}
\label{sec:cleaning-mrs}

\begin{table}[tb]
\centering\small
\begin{tabular}{llrrr}\hline
\bf Dataset                  & \bf Part    &  \bf MRs & \bf Refs & \bf SER(\%) \\\hline
\multirow{3}{*}{Original}    & \textsc{Train}       &   4,862  &  42,061  &   17.69 \\
                             & \textsc{Dev}         &     547  &   4,672  &   11.42 \\
                             & \textsc{Test}        &     630  &   4,693  &   11.49 \\\hdashline[0.5pt/2pt]
\multirow{3}{*}{Cleaned}     & \textsc{Train}       &   8,362  &  33,525  &  (0.00) \\
                             & \textsc{Dev}         &   1,132  &   4,299  &  (0.00) \\
                             & \textsc{Test}        &   1,358  &   4,693  &  (0.00) \\\hline
\end{tabular}
\caption{Data statistics comparison for the original E2E data and our cleaned version (number of distinct MRs, total number of textual references, SER as measured by our slot matching script, see Section~\ref{sec:cleaning-mrs}).}
\label{tab:cleaned-data-stats}
\end{table}

To produce a cleaned version of the E2E data, we used the original human textual references, but paired them with correctly matching MRs.\footnote{Note that this can be done automatically, unlike fixing the references to match the original MRs.} 
To this end, we reimplemented the slot matching script of \citet{reed_can_2018}, which tags MR slots and values using regular expressions.
We tuned our expressions based on the first 500 instances from the E2E development set 
and ran the script on the full dataset, producing corrected MRs for all human references (see Figure~\ref{fig:e2e-data}).
The differences against the original MRs allow us to compute the \emph{semantic/slot error rate} \cite[SER;][]{wen:emnlp2015,reed_can_2018,dusek_evaluating_2019}:
\vspace{-0.1cm}
\begin{equation*}\label{eq:ser}
\mbox{SER} = \frac{\#\mbox{added} + \#\mbox{missing} + \#\mbox{wrong value}}{\#\mbox{slots}}
\end{equation*}
To guarantee the integrity of the test set,
we removed instances from the \textsc{train} (training) and \textsc{dev} (development) sets that overlapped the \textsc{test} set. 
This resulted in 20\% reduction for \textsc{train} and ca.~8\% reduction for \textsc{dev} in terms of references (see Table~\ref{tab:cleaned-data-stats}).
On the other hand, the number of distinct MRs rose sharply after reannotation; the MRs also have more variance in the number of attributes. 
This means that the cleaned dataset is more complex overall, with fewer references per MR and more diverse MRs.

We manually evaluated 200 randomly chosen instances from the cleaned \textsc{train} set to check the accuracy of the slot matching script. 
We found that the slot matching script itself has a SER of 4.2\%, with 39 instances (19.5\%) not 100\% correctly rated. 
This is much lower than the E2E dataset authors' own manual assessment of ca.~40\% noisy instances \cite{dusek_evaluating_2019} and the script's rating of the whole dataset (mean SER: 16.37\%),
and comparable to the slot matching script of \citet{juraska_slug2slug:_2018} evaluated on the same data.\footnote{\citet{juraska_slug2slug:_2018}'s script reaches 6.2\% SER and 60 instances with errors, most of which is just omitting the \emph{eatType[restaurant]} value. If we ignore this value, it gets 1.9\% SER and 20 incorrect instances. We did not use this script as it was not available to us until very shortly before the camera-ready deadline. The script is now accessible under \url{https://github.com/jjuraska/slug2slug}. We plan to further improve our slot matching script based on errors found during the manual evaluation and comparison to \citet{juraska_slug2slug:_2018}.}

\section{Evaluating the Impact on Neural NLG}
\label{sec:evaluating-impact}

We chose two recent neural end-to-end NLG systems, which represent two
different approaches to semantic control and have been widely used and extended by the research community.  

\subsection{TGen}
\label{sec:tgen}

 TGen \cite{Dusek:ACL16} is the baseline system used in the E2E challenge.\footnote{\url{https://github.com/UFAL-DSG/tgen}} 
 TGen is in essence a vanilla sequence-to-sequence (seq2seq) model with attention \cite{bahdanau_neural_2015} using LSTM cells where input MRs are encoded as sequences of triples in the form (dialogue act, slot, value).\footnote{The dialogue act is constant/ignored for the E2E dataset since it's not part of the MRs there.} 
TGen adds to the standard seq2seq setup a reranker that selects the output with the lowest SER from the decoder output beam ($n$-best list). 
SER is estimated based on a classifier trained to identify the MR corresponding to a given text.
We use the default TGen parameters for the E2E data, experimenting with three variants:
\squishlist
\item \textbf{TGen without reranker:} a vanilla seq2seq model with attention (TGen$-$);
\item \textbf{TGen with default reranker:} the same augmented with an LSTM encoder and binary classifier for individual slot-value pairs;
\item \textbf{TGen with oracle reranker:} directly uses the slot matching script to compute SER (TGen$+$).
\squishend
We fixed the parameters of the main seq2seq generator to see the direct influence of each reranker, without the added effect of random initialization.

\begin{table*}[tb]
\footnotesize\centering
\begin{tabular}{lcl|ccccc|cccc}\hline
\bf \textsc{Train}          & \bf \textsc{Test}              
        & \bf System              & \bf BLEU & \bf NIST & \bf \hspace{-2mm}METEOR\hspace{-2mm} & \bf \hspace{-1mm}ROUGE-L\hspace{-3mm} & \bf CIDEr & \bf Add & \bf Miss & \bf Wrong &\bf SER\Tstrut \\\hline
\multirow{4}{*}{Original} & \parbox[t]{2mm}{\multirow{14}{*}{\rotatebox[origin=c]{90}{{\bf Original}}}} 
        & TGen$-$  & 63.37 & 7.7188 & 41.99 & 68.53 & 1.9355 & 00.06 & 15.77 & 00.11 & 15.94\Tstrut  \\ 
    &   & TGen     & 66.41 & 8.5565 & 45.07 & 69.17 & 2.2253 & 00.14 & 04.11 & 00.03 & 04.27  \\ 
    &   & TGen$+$  & 67.06 & 8.5871 & 45.83 & 69.73 & 2.2681 & 00.04 & 01.75 & 00.01 & 01.80  \\ 
    &   & SC-LSTM  & 39.11 & 5.6704 & 36.83 & 50.02 & 0.6045 & 02.79 & 18.90 & 09.79 & 31.51 \\\cdashline{1-1}[0.5pt/2pt]\cdashline{3-12}[0.5pt/2pt]
\multirow{4}{*}{Cleaned}  & 
        & TGen$-$  & 65.87 & 8.6400 & 44.20 & 67.51 & 2.1710 & 00.20 & 00.56 & 00.21 & 00.97\Tstrut  \\ 
    &   & TGen     & 66.24 & 8.6889 & 44.66 & 67.85 & 2.2181 & 00.10 & 00.02 & 00.00  & 00.12  \\ 
    &   & TGen$+$  & 65.97 & 8.6630 & 44.45 & 67.59 & 2.1855 & 00.02 & 00.00 & 00.00 & 00.03  \\ 
    &   & SC-LSTM  & 38.52 & 5.7125 & 37.45 & 48.50 & 0.4343 & 03.85 & 17.39 & 08.12 & 29.37 \\\cline{1-1}\cline{3-12}
\multirow{3}{*}{\shortstack{Cleaned \\ missing}} & 
        & TGen$-$  & 66.28 & 8.5202 & 43.96 & 67.83 & 2.1375 & 00.14 & 02.26 & 00.22 & 02.61\Tstrut  \\ 
    &   & TGen     & 67.00 & 8.6889 & 44.97 & 68.19 & 2.2228 & 00.06 & 00.44 & 00.03 & 00.53  \\ 
    &   & TGen$+$  & 66.74 & 8.6649 & 44.84 & 67.95 & 2.2018 & 00.00 & 00.21 & 00.03 & 00.24 \\\cdashline{1-1}[0.5pt/2pt]\cdashline{3-12}[0.5pt/2pt] 
\multirow{3}{*}{\shortstack{Cleaned \\ added}}  & 
        & TGen$-$  & 64.40 & 7.9692 & 42.81 & 68.87 & 2.0563 & 00.01 & 13.08 & 00.00 & 13.09\Tstrut  \\ 
    &   & TGen     & 66.23 & 8.5578 & 45.12 & 68.87 & 2.2548 & 00.04 & 03.04 & 00.00 & 03.09  \\ 
    &   & TGen$+$  & 65.96 & 8.5238 & 45.49 & 68.79 & 2.2456 & 00.00 & 01.44 & 00.00 & 01.45  \\\hline 
\end{tabular}
\caption{Results evaluated on the original test set (averaged over 5 runs with different random initialisation). See Section~\ref{sec:automatic} for explanation of metrics. All numbers except NIST and ROUGE-L are percentages. Note that the numbers are \emph{not} comparable to Table~\ref{tab:results-clean-testset} as the test set is different.}
\label{tab:results}
\end{table*}

\begin{table*}[tb]
\footnotesize\centering
\begin{tabular}{lcl|ccccc|cccc}\hline
\bf \textsc{Train}          & \bf\textsc{Test}              
        & \bf System\hspace{3mm}              & \bf BLEU & \bf NIST & \bf \hspace{-2mm}METEOR\hspace{-2mm} & \bf \hspace{-1mm}ROUGE-L\hspace{-3mm} & \bf CIDEr & \bf Add & \bf Miss & \bf Wrong &\bf SER\Tstrut \\\hline
\multirow{4}{*}{Original} & \parbox[t]{2mm}{\multirow{14}{*}{\rotatebox[origin=c]{90}{{\bf Cleaned}}}}
        & TGen$-$  & 36.85 & 5.3782 & 35.14 & 55.01 & 1.6016 & 00.34 & 09.81 & 00.15 & 10.31\Tstrut  \\ 
    &   & TGen     & 39.23 & 6.0217 & 36.97 & 55.52 & 1.7623 & 00.40 & 03.59 & 00.07 & 04.05  \\ 
    &   & TGen$+$  & 40.25 & 6.1448 & 37.50 & 56.19 & 1.8181 & 00.21 & 01.99 & 00.05 & 02.24  \\ 
    &   & SC-LSTM  & 23.88 & 3.9310 & 32.11 & 39.90 & 0.5036 & 07.73 & 17.76 & 09.52 & 35.03 \\\cdashline{1-1}[0.5pt/2pt]\cdashline{3-12}[0.5pt/2pt]
\multirow{4}{*}{Cleaned}  & 
        & TGen$-$  & 40.19 & 6.0543 & 37.38 & 55.88 & 1.8104 & 00.17 & 01.31 & 00.25 & 01.72\Tstrut \\ 
    &   & TGen     & 40.73 & 6.1711 & 37.76 & 56.09 & 1.8518 & 00.07 & 00.72 & 00.08 & 00.87  \\ 
    &   & TGen$+$  & 40.51 & 6.1226 & 37.61 & 55.98 & 1.8286 & 00.02 & 00.63 & 00.06 & 00.70  \\ 
    &   & SC-LSTM  & 23.66 & 3.9511 & 32.93 & 39.29 & 0.3855 & 07.89 & 15.60 & 08.44 & 31.94 \\\cline{1-1}\cline{3-12}
\multirow{3}{*}{\shortstack{Cleaned \\ missing}} & 
        & TGen$-$  & 40.48 & 6.0269 & 37.26 & 56.19 & 1.7999 & 00.43 & 02.84 & 00.26 & 03.52\Tstrut  \\ 
    &   & TGen     & 41.57 & 6.2830 & 37.99 & 56.36 & 1.8849 & 00.37 & 01.40 & 00.09 & 01.86  \\ 
    &   & TGen$+$  & 41.56 & 6.2700 & 37.94 & 56.38 & 1.8827 & 00.21 & 01.04 & 00.07 & 01.31  \\\cdashline{1-1}[0.5pt/2pt]\cdashline{3-12}[0.5pt/2pt] 
\multirow{3}{*}{\shortstack{Cleaned \\ added}}  & 
        & TGen$-$  & 35.99 & 5.0734 & 34.74 & 54.79 & 1.5259 & 00.02 & 11.58 & 00.02 & 11.62\Tstrut  \\ 
    &   & TGen     & 40.07 & 6.1243 & 37.45 & 55.81 & 1.8026 & 00.05 & 03.23 & 00.01 & 03.29  \\ 
    &   & TGen$+$  & 40.80 & 6.2197 & 37.86 & 56.13 & 1.8422 & 00.01 & 01.87 & 00.01 & 01.88  \\\hline 
\end{tabular}
\caption{Results evaluated on the cleaned test set (cf.~Table~\ref{tab:results} for column details; note that the numbers are \emph{not} comparable to Table~\ref{tab:results} as the test set is different).
}
\label{tab:results-clean-testset}
\end{table*}

\begin{table}[tb]
\vspace{-2mm}
\footnotesize\centering
\begin{tabular}{lrrrr}\hline
\bf Training data & \bf Add & \bf Miss & \bf Wrong & \bf Disfl  \\\hline
Original        & 0 & 22 & 0 & 14\Tstrut \\
Cleaned added   & 0 & 23 & 0 & 14 \\
Cleaned missing & 0 &  1 & 0 &  2 \\
Cleaned         & 0 &  0 & 0 &  5 \\\hline
\end{tabular}
\caption{Results of manual error analysis of TGen on a sample of 100 instances from the original test set: total absolute numbers of errors we found (added, missed, wrong values, slight disfluencies).}
\label{tab:manual-results}
\vspace{-3mm}
\end{table}

\subsection{SC-LSTM}
\label{sec:sclstm}

In contrast to seq2seq architecture used by TGen, the Semantically Controlled LSTM \citep[SC-LSTM,][]{wen:emnlp2015} uses a learned gating mechanism to selectively express parts of the MR during generation.
We use the SC-LSTM model provided as part of the \rnnlg{} repository\footnote{\url{https://github.com/shawnwun/RNNLG}} with minor changes to improve comparability to TGen.
Most importantly, we incorporate the tokenization and normalization used by TGen into \rnnlg.
Since the word embeddings provided with \rnnlg\ only cover about half of the tokens in the E2E dataset, we use randomly initialised word embeddings (dimension 50; same as TGen).


\section{Evaluation and Results}
\label{sec:evaluation}


To measure the effect of noisy data, we compare systems trained on the original data against systems trained using cleaned \textsc{train} and validation (=\textsc{dev}) sets; we perform the comparisons both on the original and the cleaned \textsc{test} sets.
Note that only scores on the same test set are directly comparable as the cleaned \textsc{test} set has more diverse MRs and fewer references per MR (i.e.\ numbers in Tables~\ref{tab:results} and~\ref{tab:results-clean-testset} cannot be compared across tables; cf.~Section~\ref{sec:cleaning-mrs}).


\subsection{Automatic Metrics}
\label{sec:automatic}

We use freely available \underline{w}ord-\underline{o}verlap-based evaluation \underline{m}etrics (WOM) scripts that come with the E2E data~\cite{dusek_evaluating_2019},\footnote{\url{https://github.com/tuetschek/e2e-metrics}}
 supporting BLEU \cite{papineni2002bleu}, NIST \cite{nist}, ROUGE-L \cite{lin2004rouge}, METEOR \cite{lavie_meteor:_2007} and CIDEr \cite{cider}.
In addition, we use our slot matching script for SER (cf.~Section~\ref{sec:cleaning-mrs}).
We also show detailed results for the percentages of added and missed slots and wrong slot values.\footnote{Absolute numbers of errors and number of completely correct instances are shown in Table~\ref{tab:absolute} in the Supplementary.}

The results in Table~\ref{tab:results} (top half) for the original setup confirm that the ranking mechanism for TGen is effective for both WOMs and SER, whereas the SC-LSTM seems to have trouble scaling to the E2E dataset. 
We hypothesise that this is mainly due to the amount of delexicalisation required.
However, the main improvement of SER comes from training on cleaned data with up to 97\% error reduction with the ranker and  94\% without.\footnote{$\frac{0.12}{4.27} = 0.028$ and $\frac{0.97}{15.94} = 0.061$} 
In other words, just cleaning the training data has a much more dramatic effect than just using a semantic control mechanism, such as the reranker (0.97\% vs.\ 4.27\% SER). 
WOMs are slightly lower for TGen trained on the cleaned data, except for NIST, which gives more importance to matching less frequent $n$-grams. 
This suggests better preservation of content at the expense of slightly lower fluency.

The results for testing on cleaned data (Table~\ref{tab:results-clean-testset}, top half) confirm the positive impact of cleaned training data and also show that the cleaned test data is more challenging 
(cf.\ Section \ref{sec:cleaning-mrs}), as reflected in the lower WOMs.
This raises the question whether the improved results from clean training data are due to seeing more challenging examples at training time. 
However, the improved results for training and testing on clean data (i.e.\ seeing equally challenging examples at training and test time), suggest the increase in performance can be attributed to data accuracy rather than diversity.

Looking at the detailed results for the number of added, missing, and wrong-valued slots (Add, Miss, Wrong),
 we observe more deletions than insertions, i.e.\ the models more often fail to realise part of the MR, rather than hallucinating additional information. 
 To investigate whether this effect stems from the training data, we partially cleaned the data of missing or added information only.\footnote{We only performed these experiments on TGen because of the low performance of SC-LSTM in general.} 
 However, the results in bottom halves of Tables~\ref{tab:results} and~\ref{tab:results-clean-testset} do not support our hypothesis: we observe the main effect on SER from cleaning the missed slots, reducing both insertions and deletions.
 Again, one possible explanation is that cleaning the missing slots provided more complex training examples. 

\subsection{Manual Error Analysis}

We carried out a detailed manual error analysis of selected systems to confirm the automatic metrics results, performing a blind annotation of semantic and fluency errors (not a human preference rating).
We evaluated a sample of 100 outputs on the original test set produced by TGen with the default reranker trained using all four cleaning settings (original data, cleaned missing slots, cleaned added slots, fully cleaned).
The results in Table \ref{tab:manual-results} confirm the findings of the automatic metrics: systems trained on the fully cleaned set or the set with cleaned missing slots have near-perfect performance, with the fully-cleaned one showing a few more slight disfluencies than the other. 
The systems trained on the original data or with cleaned added slots clearly perform worse in terms of both semantic accuracy and fluency.
All fluency problems we found were very slight and no added or wrong-valued slots were found, so missed slots are the main problem.

The manual error analysis also served to assess the accuracy of the SER measuring script on system outputs. Since NNLG tends to use more frequent phrasing, we expected better performance than on the dataset itself, and this proved true: 
we only found 2 errors in the 400 system outputs (i.e.\ 99.5\% of instances and 99.93\% of slots were matched correctly).
This confirms that the automatic SER numbers reflect the semantic accuracy of individual systems very closely.

\section{Discussion and Related Work}

We present a detailed study
of semantic errors in NNLG outputs and how these relate to noise in training data.
We found that even imperfectly cleaned input data significantly improves semantic accuracy for seq2seq-based generators (up to 97\% relative error reduction with the reranker), while only causing a slight decrease  in fluency.

Contemporaneous with our work is the effort of \citet{Hallucination:ACL2019}, who 
focus on automatic data cleaning using a NLU iteratively bootstrapped from the noisy data. 
Their analysis similarly finds that omissions are more common than hallucinations. 
Correcting for missing slots, 
i.e.\ forcing the generator to verbalise all slots during training,
leads to the biggest performance improvement.
This phenomenon is also observed by \citet{dusek_findings_2018,dusek_evaluating_2019} for systems in the E2E NLG challenge, but stands in contrast to work on related tasks, which mostly reports on hallucinations (i.e.\ adding information not grounded in the input), as observed for image captioning \cite{rohrbach-etal-2018-object}, sports report generation \cite{wiseman_challenges_2017}, machine translation \cite{koehn_six_2017,lee2019hallucinations}, and question answering \cite{feng_pathologies_2018}.
These previous works suggest that the most likely case of hallucinations is an over-reliance on language priors, i.e.\ memorising `which words go together'. 
Similar priors could equally exist in the E2E data for omitting a slot; this might be connected with the fact that the E2E test set MRs tend to be longer than training MRs (6.91 slots on average for test MRs vs.\ 5.52 for training MRs) and that a large part of them is `saturated', i.e. contains all possible 8 attributes.

 
 Furthermore, in accordance with our observations, related work also reports a relation between hallucinations and data diversity: \citet{rohrbach-etal-2018-object} observe an increase for ``novel compositions of objects at test time'', i.e.\ non-overlapping test and training sets (cf.\ Section \ref{sec:cleaning-mrs}); whereas \citet{lee2019hallucinations} reports data augmentation as one of the most efficient counter measures. 
 In future work, we plan to experimentally manipulate these factors to disentangle the relative contributions of data cleanliness and diversity.





\section*{Acknowledgments}

This research received funding from the EPSRC projects  DILiGENt (EP/M005429/1) and  MaDrIgAL (EP/N017536/1) and Charles University project PRIMUS/19/SCI/10. 
The authors would also like to thank Prof.~Ehud Reiter, whose blog\footnote{\url{https://ehudreiter.com/}} inspired some of this research.

\bibliographystyle{acl_natbib}
\bibliography{acl2019}

\clearpage
\appendix

\onecolumn
\large\bf Semantic Noise Matters for Neural Natural Language Generation: Supplementary
\bigskip

\begin{table*}[h]
\footnotesize\centering
\begin{tabular}{lcl|rrrr}\hline
\bf \textsc{Train}          & \bf\textsc{Test}              
        & \bf System\hspace{3mm} & \bf Add & \bf Miss & \bf Wrong &\bf InstOK\Tstrut \\\hline
\multirow{4}{*}{Original} & \parbox[t]{2mm}{\multirow{14}{*}{\rotatebox[origin=c]{90}{{\bf Original}}}} 
        & TGen$-$  &   2.8 &  686.2 &   4.8 &  192.2\Tstrut \\ 
    &   & TGen     &   6.0 &  178.8 &   1.2 &  496.4 \\ 
    &   & TGen$+$  &   1.6 &   76.2 &   0.4 &  558.2 \\ 
    &   & SC-LSTM  & 121.6 &  823.6 & 426.2 &    7.8\\ \cdashline{1-1}[0.5pt/2pt]\cdashline{3-7}[0.5pt/2pt]
\multirow{4}{*}{Cleaned}  & 
        & TGen$-$  &   8.8 &  24.2 &   9.0 &  591.6\Tstrut \\ 
    &   & TGen     &   4.2 &   0.8 &   0.2 &  624.8 \\ 
    &   & TGen$+$  &   1.0 &   0.2 &   0.2 &  628.6 \\ 
    &   & SC-LSTM  & 167.6 & 757.2 & 353.4 &   14.0 \\ \cdashline{1-1}[0.5pt/2pt]\cdashline{3-7}[0.5pt/2pt]
\multirow{3}{*}{\shortstack{Cleaned \\ missing}} & 
        & TGen$-$  &  6.0 &   98.2 &  9.4 &  525.2\Tstrut \\ 
    &   & TGen     &  2.6 &   19.0 &  1.4 &  608.0 \\ 
    &   & TGen$+$  &  0.0 &    9.0 &  1.4 &  620.6 \\\cdashline{1-1}[0.5pt/2pt]\cdashline{3-7}[0.5pt/2pt] 
\multirow{3}{*}{\shortstack{Cleaned \\ added}}  & 
        & TGen$-$  &  0.4 &  569.2 &  0.2 &  234.0\Tstrut \\ 
    &   & TGen     &  2.0 &  132.2 &  0.2 &  501.6 \\ 
    &   & TGen$+$  &  0.2 &   62.8 &  0.2 &  567.0 \\\hline 
\multirow{4}{*}{Original} & \parbox[t]{2mm}{\multirow{14}{*}{\rotatebox[origin=c]{90}{{\bf Cleaned}}}}
        & TGen$-$  &  39.4 & 1135.6 &   17.8 & 1089.4\Tstrut  \\ 
    &   & TGen     &  45.6 &  415.4 &    7.8 & 1469.8  \\ 
    &   & TGen$+$  &  23.6 &  230.2 &    5.2 & 1608.8  \\ 
    &   & SC-LSTM  & 858.6 & 1972.2 & 1057.6 &   39.0\\ \cdashline{1-1}[0.5pt/2pt]\cdashline{3-7}[0.5pt/2pt]
\multirow{4}{*}{Cleaned}  & 
        & TGen$-$  &  19.0 &  151.2 & 28.6 & 1667.8\Tstrut \\ 
    &   & TGen     &   7.8 &   83.0 &   9.6 & 1751.4 \\ 
    &   & TGen$+$  &   1.8 &   72.6 &   7.0 & 1768.8 \\ 
    &   & SC-LSTM  & 876.2 & 1732.4 & 937.4 &   78.0\\ \cdashline{1-1}[0.5pt/2pt]\cdashline{3-7}[0.5pt/2pt]
\multirow{3}{*}{\shortstack{Cleaned \\ missing}} & 
        & TGen$-$  & 49.4 &  328.4 & 30.0 & 1482.6\Tstrut  \\ 
    &   & TGen     & 42.8 &  162.0 & 10.8 & 1643.2 \\ 
    &   & TGen$+$  & 24.0 &  120.0 &  8.0 & 1702.8  \\\cdashline{1-1}[0.5pt/2pt]\cdashline{3-7}[0.5pt/2pt] 
\multirow{3}{*}{\shortstack{Cleaned \\ added}}  & 
        & TGen$-$  &  2.2 & 1340.2 &  2.8 &  959.8\Tstrut \\ 
    &   & TGen     &  6.0 &  373.6 &  1.8 & 1518.6 \\ 
    &   & TGen$+$  &  0.8 &  216.6 &  0.8 & 1646.2 \\\hline 
\end{tabular}
\caption{Absolute numbers of errors (added slots/missed slots/wrong slot values) and numbers of completely correct instances in all our experiments (compare to Tables~\ref{tab:results} and~\ref{tab:results-clean-testset} in the paper). Note that (1) the numbers are averages over 5 runs with different random network initializations, hence the non-integer values; (2) only numbers in the top half and the bottom half (with the same test set) are comparable. The original test set has 630 MRs and 4,352 slots in total. The cleaned test set has 1,847 MRs and 11,547 slots; however, for the runs with SC-LSTM these counts are 1,800 and 11,101, respectively, since some items had to be dropped due to preprocessing issues.}
\label{tab:absolute}
\end{table*}

\end{document}